\documentclass[letterpaper, 10 pt, journal, twoside]{IEEEtran}
\makeatletter
\let\NAT@parse\undefined
\makeatother
\usepackage[colorlinks]{hyperref}
\hypersetup{colorlinks=true,linkcolor=red,citecolor=blue,urlcolor=magenta}

\usepackage[utf8]{inputenc}
\usepackage[T1]{fontenc}
\usepackage{graphicx}
\usepackage{multirow}
\usepackage{verbatim}
\usepackage{makecell}
\usepackage[table,dvipsnames]{xcolor} 
\usepackage{amsmath} 

\begin{document}

\title{\Large\bf Bi-Mapper: Holistic BEV Semantic Mapping for Autonomous Driving}

\author{Siyu Li$^{1}$, Kailun Yang$^{1}$, Hao Shi$^{2}$, Jiaming Zhang$^{3,4}$, Jiacheng Lin$^{5}$, Zhifeng Teng$^{3}$, and Zhiyong Li$^{1,5}$
\thanks{Manuscript received: May 6, 2023; Revised: July 4, 2023; Accepted: August 28, 2023.}
\thanks{This paper was recommended for publication by Editor Javier Civera upon evaluation of the Associate Editor and Reviewers’ comments.}
\thanks{This work was supported in part by the National Natural Science Foundation of China (No.U21A20518 and No.61976086) and in part by Hangzhou SurImage Technology Company Ltd. \textit{(Corresponding authors: Kailun Yang and Zhiyong Li.)}}
\thanks{$^{1}$S. Li, K. Yang, and Z. Li are with the School of Robotics, Hunan University, Changsha 410082, China (email: kailun.yang@hnu.edu.cn; zhiyong.li@hnu.edu.cn).}%
\thanks{$^{2}$H. Shi is with the State Key Laboratory of Modern Optical Instrumentation, Zhejiang University, Hangzhou 310027, China.}%
\thanks{$^{3}$J. Zhang and Z. Teng are with the Institute for Anthropomatics and Robotics, Karlsruhe Institute of Technology, Karlsruhe 76131, Germany.}%
\thanks{$^{4}$J. Zhang is also with the Department of Engineering Science, University of Oxford, Oxford OX1 3PJ, UK.}%
\thanks{$^{5}$J. Lin and Z. Li are with the School of Computer Science and Electronic Engineering, Hunan University, Changsha 410082, China.}
\thanks{Digital Object Identifier (DOI): see top of this page.}
}
\markboth{IEEE Robotics and Automation Letters. Preprint Version. Accepted August, 2023}
{Li \MakeLowercase{\textit{et al.}}: Bi-Mapper: Holistic BEV Semantic Mapping for Autonomous Driving} 

\maketitle

\begin{abstract}
A semantic map of the road scene, covering fundamental road elements, is an essential ingredient in autonomous driving systems. It provides important perception foundations for positioning and planning when rendered in the Bird's-Eye-View (BEV). Currently, the prior knowledge of hypothetical depth can guide the learning of translating front perspective views into BEV directly with the help of calibration parameters. However, it suffers from geometric distortions in the representation of distant objects. In addition, another stream of methods without prior knowledge can learn the transformation between front perspective views and BEV implicitly with a global view. Considering that the fusion of different learning methods may bring surprising beneficial effects, we propose a Bi-Mapper framework for top-down road-scene semantic understanding, which incorporates a global view and local prior knowledge. To enhance reliable interaction between them, an asynchronous mutual learning strategy is proposed. At the same time, an \emph{Across-Space Loss (ASL)} is designed to mitigate the negative impact of geometric distortions. Extensive results on nuScenes and Cam2BEV datasets verify the consistent effectiveness of each module in the proposed Bi-Mapper framework. Compared with exiting road mapping networks, the proposed Bi-Mapper achieves $2.1\%$ higher IoU on the nuScenes dataset. Moreover, we verify the generalization performance of Bi-Mapper in a real-world driving scenario. The source code is publicly available at \href{https://github.com/lynn-yu/Bi-Mapper}{BiMapper}.

\end{abstract}
\section{Introduction}
\IEEEPARstart{I}{n} autonomous driving systems, a semantic map is an important basic element, which affects the downstream working, including location and planning.
Recently, the Bird' s-Eye-View (BEV) map has shown an outstanding performance~\cite{14}.
It can construct a map as the simple paradigm of a High-Definition map (HD-map), on which the path planning can be easily generated~\cite{1}.
Similar to the HD map representation, a BEV road map can represent basic road elements, such as road dividers, pedestrian crossing, and boundaries, which enable holistic driving scene understanding~\cite{peng2022mass}.

Nowadays, a BEV road map is usually transformed from front-view images~\cite{can2022understanding_road_semantics}.
It can be directly constructed via depth information, according to the intrinsic and extrinsic parameters.
Yet, consumer-grade monocular cameras without depth information can provide a cost-effective choice.
Therefore, the core of BEV mapping is to effectively learn high-quality features and predict top-down semantics from front-view scenes.

\begin{figure}[tb]
      \centering
      \includegraphics[scale=0.5]{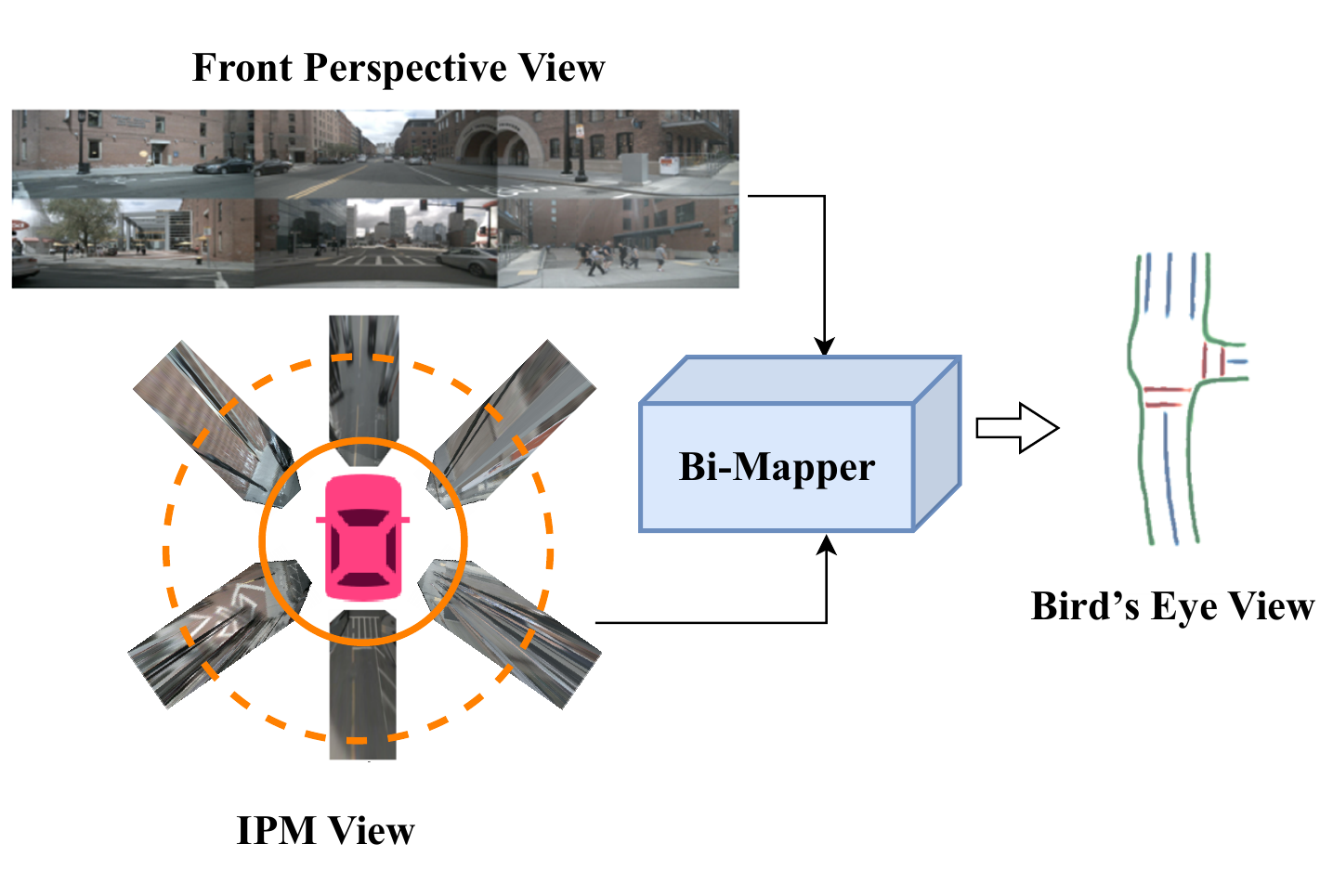}
      \vspace{-0.55cm}
      \caption{Bi-Mapper constructs a BEV road map from the front perspective view and the IPM view. They describe the relationship between objects from two different perspectives. The IPM view, produced by a hypothetical depth, has robust representations for near objects but has distortions for the far objects, shown in orange circles.}
      \label{Fig.1}
      \vspace{-0.75cm}
\end{figure}

Currently, there are two mainstream directions to fulfill BEV semantic mapping.
BEV mapping with explicit depth estimation is a direction~\cite{8}, which projects 2D pixels to 3D points based on the calibration parameters.
Inverse Perspective Mapping (IPM)~\cite{33} is one of the special cases.
It assigns a hypothetical depth to each pixel.
This prior knowledge, which has significant guidance, may bring some problems, such as geometric distortions in the representation of distant objects~\cite{2014Recent}.
As shown in Fig.~\ref{Fig.1}, far objects appear to be blurry and the near ones are relatively clear in the IPM-view images. 
Another direction relies on deep learning of a front-view transformer that implicitly learns depth information and calibration parameters~\cite{19,20}.
While learning slowly in the early period, it performs preferably after complete training. 
As both strategies have their strengths, we raise the appealing issue and ask if interactions between the two could produce complementary effects to boost BEV scene understanding.

To this end, we propose Bi-Mapper, a framework that considers prior knowledge and view-transformer learning in a parallel manner. This is the first time, to the best of our knowledge, a dual-stream framework using both perspective and IPM views, is proposed for BEV semantic mapping.
Specifically, it combines the capability from two streams, \textit{i.e.}, Local-self View stream (LV) and Global-cross View stream (GV).
The LV is under the guidance of prior knowledge of a hypothetical depth, whereas the GV leverages the self-learning capacity of the model to implicitly infer the semantics.
On the one hand, prior knowledge has its inherent guidance, which means that it can point the learning direction at the beginning of training.
Facing the negative impacts of geometric distortions, we propose an \emph{Across Space Loss} \emph{(ASL)} to alleviate them. 
It supervises the learning in a different space, namely the camera system coordinate.
Note that the ground truth for BEV is in an ego-system coordinate.
On the other hand, the information from the learning for GV is relatively scarce in the early learning phase.
Therefore, we propose an \emph{Asynchronous Mutual Learning} \emph{(AML)} strategy that starts mutual learning until both streams have equal status.
Concretely, in the beginning, only LV serves as the teacher.
The proposed asynchronous strategy is beneficial for the streams to learn effective knowledge.  

We conduct extensive experiments on nuScenes~\cite{31} and Cam2BEV~\cite{32} datasets.
Bi-Mapper achieves $37.9\%$ in IoU on the nuScenes dataset, which is $2.1\%$ higher than the best performance of existing methods.
It has an outstanding result on the Cam2BEV data with $86.7\%$ in IoU, which is $4.1\%$ higher than contemporary methods.
What is more, the proposed approach is consistently effective for bringing complementary benefits to state-of-the-art solutions, such as HDMapNet~\cite{19} and LSS~\cite{8}. It reaches an improvement of $4.0\%$ and $6.6\%$ in IoU, respectively.

In summary, the main contributions of our research lie in:
\begin{itemize}
    \item An end-to-end Bi-Mapper framework, which learns a BEV map from different points of view simultaneously, is proposed. One stream focuses on prior knowledge, whereas the other leverages the self-learning capacity, in which the complementary knowledge can be harvested via cross-stream interactions.
    \item Motivated by that prior knowledge can guide the directions of learning, an asynchronous mutual learning strategy is designed to balance two streams and alleviate the gap of prior knowledge in BEV semantic mapping.
    \item To improve the accuracy of the network, an Across-Space Loss is proposed, which can also alleviate the geometric distortion problem.
    \item An extensive set of experiments demonstrates the superiority of the proposed algorithm and the effectiveness of the key components.
\end{itemize}

\begin{figure*}[tb]
      \centering
      \includegraphics[width=0.8\linewidth]{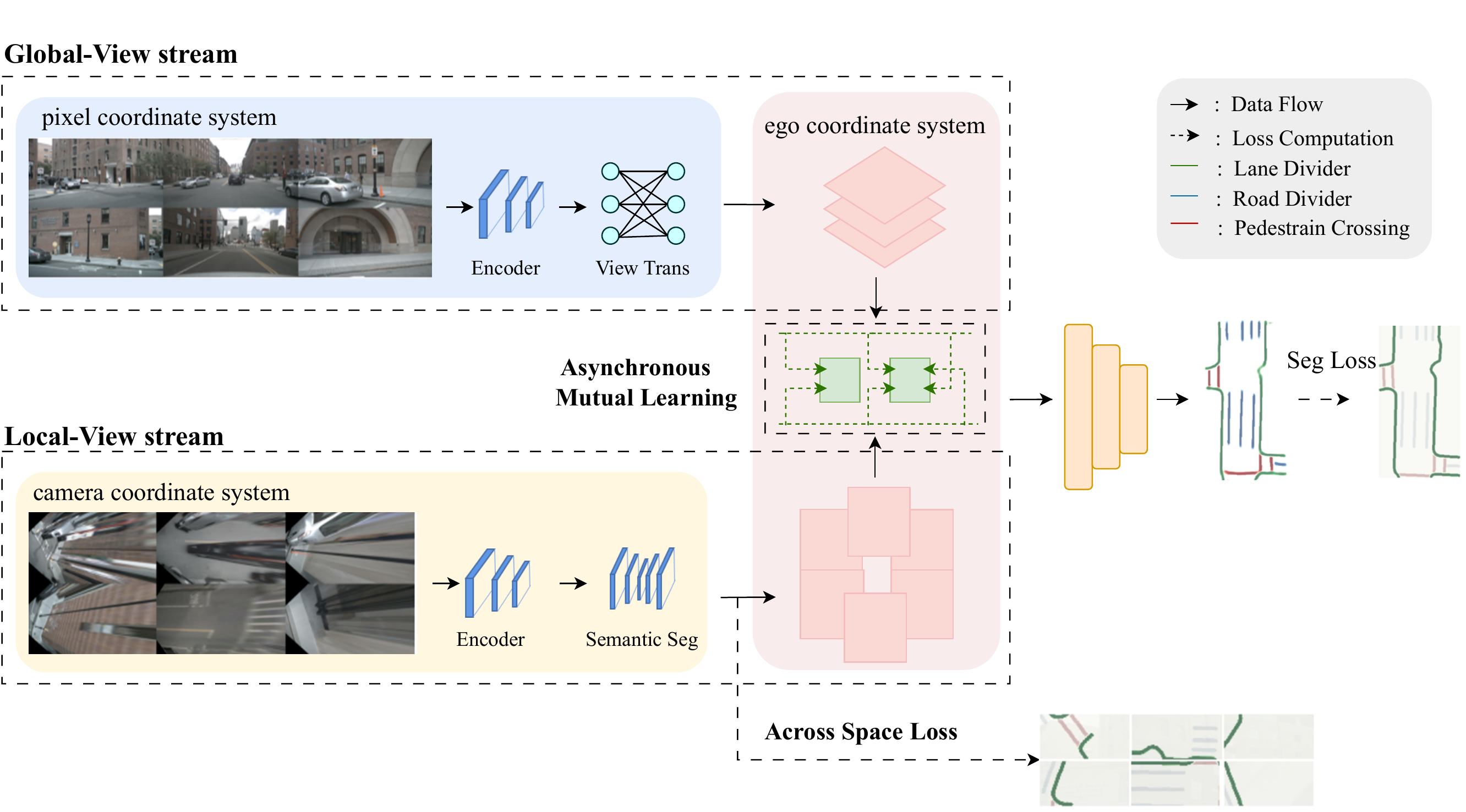}
      \caption{The framework of the proposed Bi-Mapper network. It is comprised of a global-cross view stream, a local-self view stream, asynchronous mutual learning, and a decoder module. \emph{View Trans} translates features from the pixel layer into the camera coordinate system for each view. \emph{Semantic Seg} learns semantic features for each IPM view. In addition to the segmentation loss, there is an across-space loss, which can alleviate the geometric distortion problem. }
      \label{Fig.2}
      \vspace{-0.65cm}
\end{figure*}
\section{Related Work}
The research on BEV perception has attracted scholars' attention~\cite{1,2}. In this part, we briefly outline the research progress of the BEV perception. Current research works are mainly divided into three categories: one is based on the Inverse Perspective Mapping (IPM) transformer, the second is via depth estimation, and the other leverages deep neural networks to learn this parameter model directly.

\textbf{IPM-based Methods:} IPM is the earliest and simplest solution~\cite{3,4} to achieve BEV perception.
There are many research works to learn semantic knowledge on this basis.
The work~\cite{5} first converted features from front views to top-down views through IPM and then used the network to learn semantic features. 
IPM has a significant problem in that the geometrical shape of an object may be warped in the distance.
Zou~\textit{et al.}~\cite{6} designed two branches to learn the difference between IPM and the geometric feature, which is realized via mutual learning mechanisms.
Sharing a similar motivation, we propose an idea for asynchronous learning. 
Considering the issue of inconsistent learning levels at the beginning, our method initially has only a stream as a teacher. Then mutual learning starts when two streams have a comparable level of learning, which can ensure the high reliability of the learned feature.
Moreover, Can~\textit{et al.}~\cite{7} considered the aggregation of temporal information, which can resolve the problem of dynamic obstacles. 
In addition to a semantic map we study, a vector map is also an important perception task that concentrates on vectorial instances. VectorMapNet~\cite{vecmap}, as a pioneer work about vector maps, also leveraged IPM for view transformation. 

\textbf{Depth-estimation-based Methods:}
The second paradigm addresses BEV perception via depth estimation.
The classic research work in this direction is~\cite{8}.
Philion~\textit{et al.}~estimated the probability of depth for each pixel via depth discretization.
Next, the estimated depth was used to obtain a frustum feature map that showed a 3D grid feature.
Based on this, Xie~\textit{et al.}~\cite{9} improved this method to reduce the running memory.
They assumed that the depth is evenly distributed along the light.
It means that the grid along the camera light is filled with the same feature.
Recently, 
Huang~\textit{et al.}~\cite{10} enhanced the method~\cite{9} by an optimized view transformation to speed up the inference and a data augmentation strategy.
Hu~\textit{et al.}~\cite{11} proposed an end-to-end network for perception, prediction, and planning.
They referred to the method in~\cite{8} to learn BEV mapping.
However, all these methods are based on implicitly learned depth. In~\cite{12}, it was shown that using the ground truth of depth as explicit supervision has an excellent result.
In a dynamic environment, the temporal information is significant.
Thus, Hu~\textit{et al.}~\cite{13} fused temporal features to predict current- and future states on the basis of estimated depth.
Relative to the depth source camera coordinate system, the height corresponds to the vehicle coordinate system.
Thus, the height of 3D space is worth to be studied to construct BEV maps.
Li~\textit{et al.}~\cite{14} considered the parallelism of height and time and introduced the temporal feature. Further, this improvement has been recently applied in a vector map network MapTR~\cite{maptr}.

\textbf{Learning-based Methods:}
Lu~\textit{et al.}~\cite{15} used a variational auto-encoder to learn the transformation between front-view and top-view data.
The input is an image and the output is a semantic occupancy grid map. In~\cite{8}, the depth was divided into $1m$ cells, while Roddick~\textit{et al.}~\cite{16} divided it into four parts according to the principle from near to far.
The deep feature in the pyramid was used to learn the near feature map by collapsing to a bottleneck and flattening to a feature map.
The reason was that small objects in the distance are usually not clearly shown in the deep layer.
In~\cite{17}, it was proposed a vertical transformer and a flat transformer to prove accuracy.
Pan~\textit{et al.}~\cite{18} proposed a different method that uses Multi Layers Perception (MLP) to learn the transformation.
At the same time, they applied domain adaptation for deployment in real environments.

On the knowledge of the above works, modified methods were presented.
In~\cite{19}, multi-view images were translated into the camera coordinate system via multi MLPs.
Then a BEV map can be constructed in the ego coordinate system by extrinsic parameters. Yet, we utilize MLPs to learn intrinsic and extrinsic parameters directly. IPM produced by these parameters can guide the directions of MLPs.
To enhance the accuracy, the adversarial learning for \emph{car} and \emph{road} was studied in~\cite{20}.
At the same time, they introduced a cycled self-supervision scheme. Specifically, they projected the BEV view to the front view and calculated a cycle loss.
On the basis of it, Yang~\cite{21} added a dual-attention module.
With the excellent performance of vision transformers, researchers found that it is suitable for bridging the gap between front views and BEV~\cite{22,23,24}.
However, the computation complexities of vision transformers are huge. Actually, each vertical line in an image is related to its associated ray in BEV.
Thus, the correspondence was learned via the attention mechanism in~\cite{25}. Simultaneously, the context between rays in BEV was added to enhance spatial relationships.
Gong~\textit{et al.}~\cite{26} also paid attention to this correspondence by adding geometric prior knowledge. 

Deep mutual learning, as a branch of model distillation, was introduced in~\cite{27}. The core of it is how to learn the relation between the deep features from multi-streams.
Kim~\textit{et al.}~\cite{28} demonstrated that mutual knowledge distillation not only performs satisfactorily in fusion classifiers but is also beneficial for sub-networks observably.
Thus, it has been popular in multi-modal networks.
The works~\cite{29} and~\cite{30} both focused on feature distillation between the streams of the camera and LiDAR data. The mutual learning in the former aimed at the weighted loss of features from the object box.
The latter used the mean square error between BEV feature maps from the streams of the camera and LiDAR via a mutual loss.
Unlike these works aiming at final objects, we propose an asynchronous mutual learning strategy for the feature layers that consider the learning degree of two streams. Both feature layers are fused to learn the BEV layout. 
\begin{figure}[tb]
      \centering
      \includegraphics[width=\linewidth]{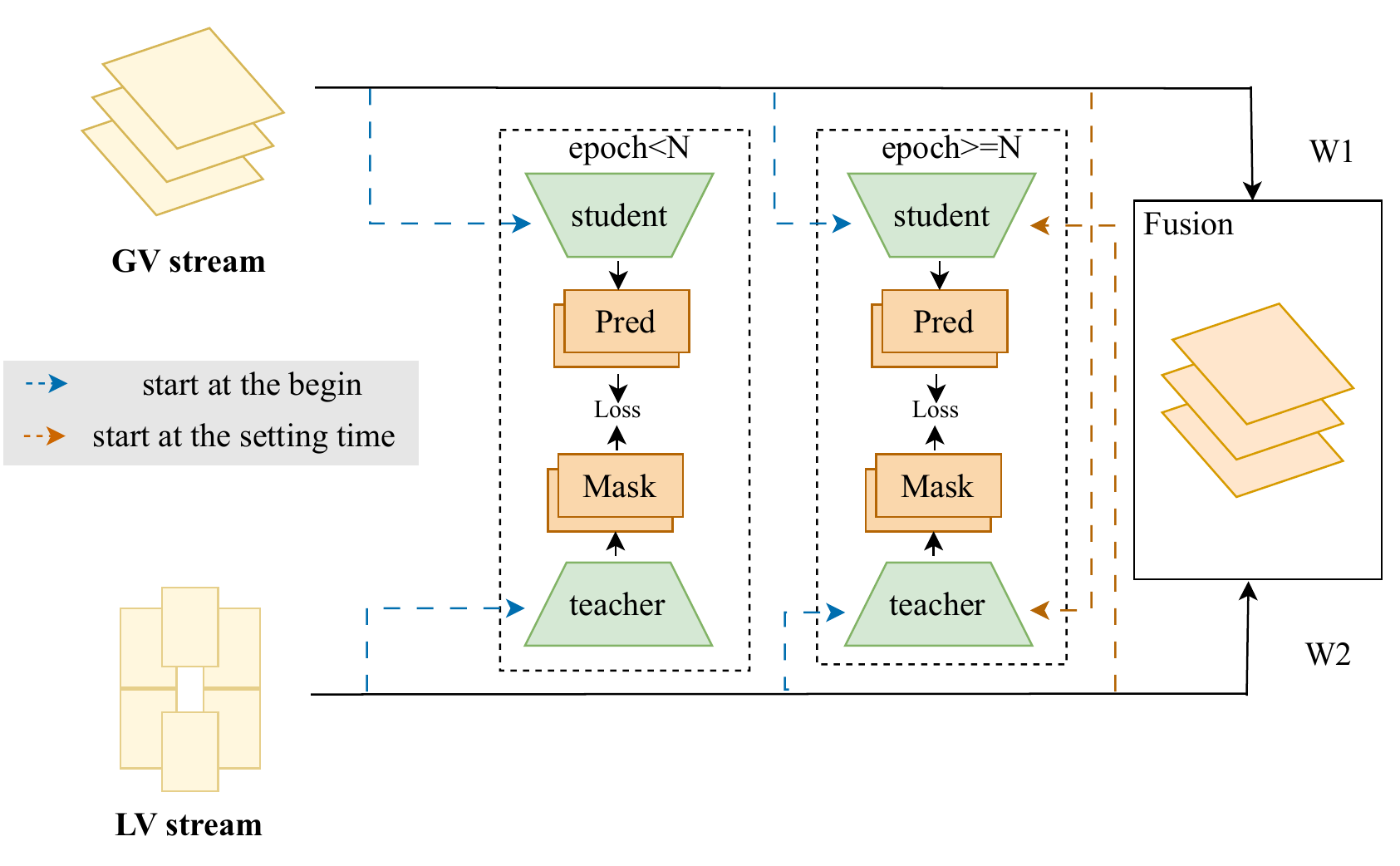}
      \vspace{-0.65cm}
      \caption{The asynchronous mutual learning and fusion module.}
      \label{Fig.3}
      \vspace{-0.75cm}
\end{figure}
\section{Method}
We first introduce three coordinate systems (Sec.~\ref{sec:problem_form}) and describe the overview of the proposed Bi-Mapper framework (Sec.~\ref{sec:bi-mapper}). Then we introduce the details of two streams of global-cross view (Sec.~\ref{sec:global_view}) and local-self view (Sec.~\ref{sec:local_view}), respectively, an across-space loss (Sec.~\ref{sec:across_loss}), and an asynchronous mutual learning module (Sec.~\ref{sec:mutual_learning}).

\begin{figure*}[tb]
      \centering
      \includegraphics[width=\linewidth]{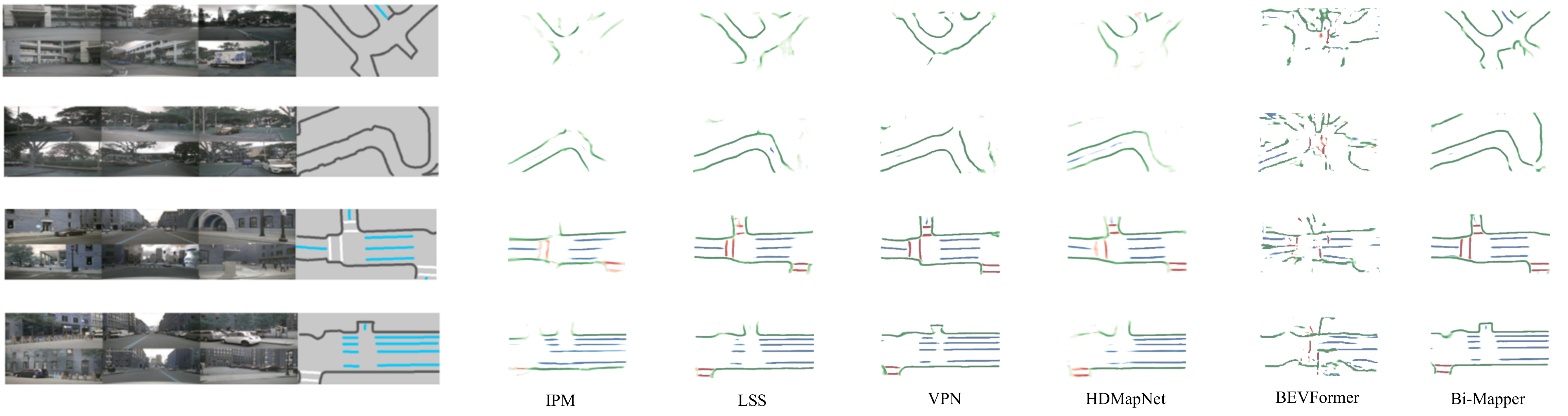}
      \vspace{-0.5cm}
      \caption{Mapping results on the nuScenes validation set. From left to right are the results of IPM~\cite{33}, LSS~\cite{8}, VPN~\cite{18}, HDMapNet~\cite{19}, BEVFormer~\cite{14}, and our proposed Bi-Mapper.}
      \label{Fig.4}
      \vspace{-0.48cm}
\end{figure*}
\begin{table*}[htb]
\renewcommand{\arraystretch}{1.1}
\caption{Comparison results on the nuScenes dataset. $*$ represents that the data is from the corresponding reference (IoU (\%)) (CD (m)).}
\vspace{-0.45cm} 
\label{table12}
\begin{center}
\begin{tabular}{cccccccccccccc}
\hline
\multirow{2}{*}{Method} & \multicolumn{4}{c}{Divider}                      & \multicolumn{4}{c}{Ped Crossing}                 & \multicolumn{4}{c}{Boundary}   & \multicolumn{1}{c}{All Classes}                   \\ \cline{2-14} 
               &IoU         & CD\_p          & CD\_L          & CD        &IoU      & CD\_p          & CD\_L           & CD       &IoU      & CD\_p          & CD\_L          & CD      &IoU       \\ \hline\hline
IPM~\cite{33}             &38.6*        & 1.045          & 0.812          & 0.941      &19.3*    & 1.085          & 1.420          & 1.232     &39.3*     & 0.523          & 1.494          & 0.968       &32.4*   \\
LSS~\cite{8}                 &38.3*    & 1.054          & 0.468          & 0.782    &14.9*      & 0.657          & 0.454          & 0.573     &39.3*     & 0.453          & 0.515          & 0.478       &30.8*   \\
VPN~\cite{18}              &36.5*       & 0.994          & 0.246          & 0.644     &15.8*     & 0.474          & 0.138          & 0.329     &35.6*     & 0.620          & 0.551          & 0.588      &29.3*    \\
HDMapNet~\cite{19}       &40.6*         & 0.874          & 0.312          & 0.699     &18.7*     & 0.358          & 1.055          & 0.946     &39.5*     & 0.285          & 0.346          & 0.323     &32.9*     \\
BEVFormer~\cite{14}      & 42.1      & 0.487         &\textbf{0.006}        &  0.212    &  23.8     & 0.201          & 0.027          & 0.118    &41.6     &0.034          & \textbf{0.001}          & 0.019     &35.8     \\
\rowcolor{gray!10}Bi-Mapper         &\textbf{43.8}           & \textbf{0.147} & 0.010 & \textbf{0.084}  &\textbf{25.7} & \textbf{0.045} & \textbf{0.009} & \textbf{0.026} &\textbf{44.2} & \textbf{0.004} & 0.006 & \textbf{0.005}  &\textbf{37.9}\\ \hline
\end{tabular}
\end{center}
\vspace{-0.95cm} 
\end{table*}

\subsection{Problem Formulation}
\label{sec:problem_form}
The transformation involves three coordinate systems, pixel, camera, and ego coordinate systems. The three coordinate systems have two transformation matrixes, an intrinsic matrix $T_{in}$ and an extrinsic matrix $T_{ex}$. 
\begin{flalign*}
    && \ T_{in}=\left[
    \begin{array}{lll}
    f_{x} & 0 & c_x \\
	0 & f_{y} & c_y \\
    0 & 0 & 1\\
    \end{array}
    \right],
    & \ &
    T_{ex}=[R|T],&&
    \vspace{-0.15cm}
\end{flalign*}
where $f_x = \alpha f, f_y = \beta f$. $f$ is the focal length. Objects in the pixel coordinate system are scaled $\alpha$ times on the u-axis and $\beta$ times on the v-axis relative to the camera coordinate system. u and v is the pixel coordinate system. Meanwhile, the origin has shifted $c_x$ and $c_y$, respectively. $R$ is the rotation matrix and $T$ is the translation matrix, which are derived from the calibration between the camera and the ego.

The transformation of three coordinate systems is computed by Eq.~\ref{eq1}. The work~\cite{19} utilizes MLPs to learn $T_{in}$ and $Z_c$. Then $T_{ex}$ is leveraged to obtain a BEV feature map in the ego coordinate system. Different from~\cite{19}, the proposed method learns $Z_c$, $T_{in}$ and $T_{ex}$ directly.
\begin{equation}
    Z_c * \left[\begin{array}{l} u \\ v \\ 1\\ \end{array} \right] = T_{in} * \left[\begin{array}{l} X_c \\ Y_c \\ Z_c\\ \end{array} \right]
    \left[\begin{array}{l} X_c \\ Y_c \\ Z_c\\ 1\\ \end{array} \right] = T_{ex} * \left[\begin{array}{l} X_w \\ Y_w \\ Z_w\\ 1\\ \end{array} \right],
    \vspace{-0.15cm}
    \label{eq1}
\end{equation}
where $X_c$, $Y_c$, and $Z_c$ are in the camera coordinate system, and $X_w$, $Y_w$, and $Z_w$ are in the ego coordinate system.

\subsection{Bi-Mapper Framework}
\label{sec:bi-mapper}

As shown in Fig.~\ref{Fig.2}, the Bi-Mapper framework includes two streams, namely the Global-cross View stream (GV) and Local-self View stream (LV), an across-space loss, and an asynchronous mutual learning module. The GV stream merges different views' features in the ego coordinate system in a manner that relies entirely on learning, which can provide results of multi-views cross-learning. We note that the physical process of transforming from pixel coordinates to BEV coordinates has barely been explored in previous learning-based work. Concretely, the BEV road map is in the ego coordinate system, the transformation between pixel and camera coordinate systems can be obtained from intrinsic parameters on the basis of a simple assumption, and the other transformation is from extrinsic parameters. Considering that the physical model of imaging is an important prior knowledge, which involves three coordinate systems -- pixel, camera, and ego, the LV stream follows this model to learn local features from multi-views, respectively. However, this simple assumption results in a geometric distortion. Therefore, we further design an Across-Space loss to reduce this side effect. The goal of two streams where the way of learning is different is concordant. For this reason, an asynchronous mutual learning module is designed to improve the ability of segmentation.
Afterward, GV and LV streams are fused into global feature maps. Finally, a segmentation head is used to obtain a BEV road map. 

\subsection{Global-Cross View}
\label{sec:global_view}

The previous work~\cite{19} has proved that MLP has the ability to construct a model between the pixel and the camera coordinate system.
The transformation between the camera coordinate system and the ego coordinate system follows extrinsic parameters.
However, this section sheds new light on it, as we employ MLP to directly transform between the pixel coordinate system and the ego coordinate system.
This can be attributed to the association of intrinsic and extrinsic parameters involving depth.
For example, depth determines their association which can be learned at the pixel level. Thus, we use MLP to learn the view transformation
between the three coordinate systems directly.
The feature maps of different views are fused into a global map. 

Specifically, images from each view are embedded by an encoder to obtain feature maps in the pixel coordinate.
Then MLP is used to translate feature maps in the pixel layer into the ego coordinate system. Different views have a shared two-level MLP architecture yet respective parameters.
Feature maps from different views are then added to a global feature map in the ego coordinate system, which can adaptively adjust the weights of cross-view features in the ego view via independent MLP layers:
\vspace{-0.15cm}
\begin{equation}
F_{g}=\sum_{n=0}^{5} \varphi_{2}^{n}(ReLU(\varphi_{1}^{n}(F_{p}^{n}))),
\vspace{-0.15cm}
\end{equation}
where $F_{g} \in R^{h \times w \times c}$ is a global feature map, $ F_{p}^{n} \in R^{H \times W \times C}$ is feature maps of $n$-th view in the pixel layer, $\varphi_{i}(\cdot) $ is the $i$-th layer MLP.
$ReLU$ denotes the rectified linear unit activation.

\subsection{Local-Self View}
\label{sec:local_view}

As the GV stream fully leverages the network to learn the transformation model, our LV stream is based on prior parameters and integrates prior knowledge into the neural network architecture. 
The aforementioned elements, including road divide, lane divide, and pedestrian crossing, serve as the research scope of this study. These elements are situated on the lane, and as such, it can be inferred that the IPM~\cite{33}, is applicable to them. As a result, the image of a plane, approximately one meter in height, within the camera coordinate system can be obtained through the utilization of intrinsic parameters. It should be noted that this image is subject to a geometric distortion, which will be analyzed in the next section (Sec.~\ref{sec:across_loss}).

To obtain semantic maps in each camera coordinate system, images from each view are first translated into their respective camera coordinate systems.
Subsequently, a basic semantic segmentation network, U-Net, is employed to generate semantic maps in each camera coordinate system. Finally, the semantic maps obtained from each camera coordinate system are merged into a global semantic map in the ego coordinate system via the use of extrinsic parameters.

\begin{figure}[tb]
    
    \setlength\tabcolsep{1pt}
    {
    \newcolumntype{P}[1]{>{\centering\arraybackslash\fontsize{7.5}{20}\selectfont}p{#1}}

    \begin{tabular}{@{\hspace{76pt}}*{4}{P{0.22\columnwidth}}@{}}
    
    {\cellcolor[rgb]{0.42,0.51,0.65}}\textcolor{white}{Road} 
    &{\cellcolor[rgb]{0.64,0.40,0.43}}\textcolor{white}{Dynamic objects}
    &{\cellcolor[rgb]{0.4,0.56,0.46}}\textcolor{white}{Static objects}  \\
    \end{tabular}
    }
    
    \centering
      \includegraphics[width=\linewidth]{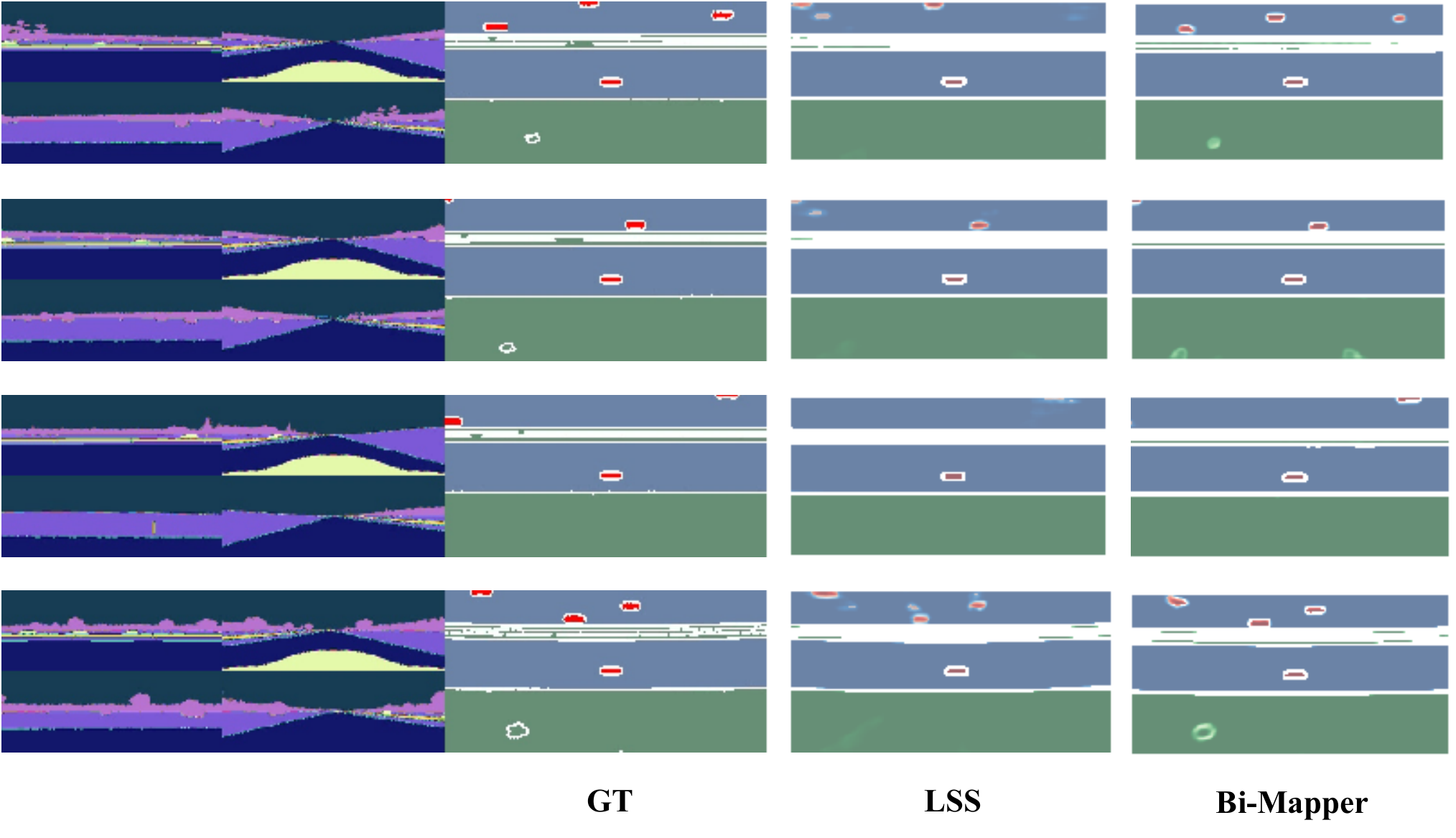}
      \vspace{-0.55cm}
      \caption{Mapping results on the Cam2BEV dataset. From left to right are the results of  LSS~\cite{8} and Bi-Mapper.}
      \label{Fig.6}
      \vspace{-0.75cm}
\end{figure}

\subsection{Across-Space Loss}
\label{sec:across_loss}

Within the process of generating a BEV road map, three coordinate systems are employed. Conventionally, the ground truth is situated in either an ego or a pixel coordinate system. 
Fewer BEV studies~\cite{19,5} have considered the ground truth in the camera coordinate system. 
But for the local-self view stream, the geometric distortions, if left unchecked, would be carried into the fusion process, ultimately hindering the learning of the other stream.
In order to mitigate these issues, we propose the utilization of an Across-Space Loss.
Concretely, the ground truth in a camera coordinate system corresponds to each view for a frame, which is derived from the ego pose and extrinsic parameters in a global map.
The difference between the results obtained from each camera coordinate system and the ground truth is calculated using cross-entropy loss, which serves as a means of ensuring the accuracy of the generated BEV road map and reducing the impact of any geometric distortions that may occur:
\vspace{-0.2cm}
\begin{equation}
    Loss_{ASL} = \sum_{j=0}^{5} L_{ij}(F_{gt},F_{pre}),
    \vspace{-0.2cm}
\end{equation}
where $j$ is the number of view and $L_{ij}$ is cross-entropy loss.

\begin{table}[tb]
\large
\renewcommand{\arraystretch}{1.1}
\caption{The comparison results on the Cam2BEV dataset.}
\vskip -2ex
\label{table8}
\begin{center}
\scalebox{0.65}{
\begin{tabular}{c|ccc|c}
\hline
Method    & Road       & Dynamic objects & Static objects      & All classes   \\ \hline \hline
IPM~\cite{33}      & 39.1          & 33.7                & 56.1          & 43.0          \\
LSS~\cite{8}      & 97.1         & 56.5                & 94.3          & 82.6          \\
VPN~\cite{18}      & 52.4          & 31.0                & 67.6          & 50.3          \\
HDMapNet~\cite{19} & 39.6          & 1.0                & 66.1          & 35.6          \\
\rowcolor{gray!10} Bi-Mapper      & \textbf{97.7} & \textbf{67.5}       & \textbf{95.0} & \textbf{86.7} \\ \hline
\end{tabular}}
\end{center}
\vspace{-0.75cm} 
\end{table}

\subsection{Asynchronous Mutual Learning and Fusion}
\label{sec:mutual_learning}
While two streams learn a BEV road map through distinct methods, they share a similar semantic expression. Moreover, they can leverage information from one another in a mutual learning setting. By default, Deep Mutual Learning (DML) is a suitable solution for this scenario. In general, DML is applied between two streams simultaneously, as long as they are supervised by the ground truth. However, in this case, GV and LV are not directly supervised by the label, necessitating the need for an asynchronous mutual learning and fusion module, as depicted in Fig.~\ref{Fig.3}.

Initially, the LV stream serves as a teacher for the GV stream due to its prior knowledge and having been supervised by the Across-Space Loss in the camera coordinate system. Conversely, the GV stream has a lower degree of learning at the beginning of training. As the GV stream learns transformation, it can teach the LV stream simultaneously through asynchronous mutual learning, which is constrained by cross-entropy loss:
\vspace{-0.15cm}
\begin{equation}
\label{eq:mutual_loss}
Loss_{mutual}=\left\{\begin{array}{ll}
    Loss_{LV}&n \textless N, \\
	Loss_{LV} + Loss_{GV} & else. \\
    \end{array}\right\}
    \vspace{-0.15cm}
\end{equation}
where $Loss_{LV}$ represents that the teacher is the LV stream. $Loss_{GV}$ represents that the teacher is the GV stream.
$N$ is the epoch of training.
In essence, the two streams learn from each other about their feature learning outcomes by predicting the foreground and background of the feature map, with the teacher being a mask.
Following mutual learning, the two streams are fused into global feature maps through a weighted addition operation.

\subsection{Training Loss}
\label{sec:loss}

Overall, the loss can be computed by:
\vspace{-0.15cm}
\begin{equation}
    Loss = Loss_{BCE} + Loss_{ASL} + \alpha \cdot Loss_{mutual}
    \vspace{-0.15cm}
\end{equation}
where $Loss_{BCE}$ is a binary cross-entropy loss that narrows the gap between prediction and the ground truth in the ego coordinate system. $Loss_{ASL}$ is the Across-Space Loss. $Loss_{mutual}$ is the result of mutual learning. $\alpha$ is a balanced weight, which is empirically set as $0.1$.

\section{Experiments}
To verify our proposed method, a series of experiments are conducted. It includes three parts. One is a comparison of state-of-the-art BEV methods. The other is an ablation study to analyze the components of Bi-Mapper. At last, we evaluate Bi-Mapper in a real robotic navigation scenario. 
\begin{table}[tb]
\large
\renewcommand{\arraystretch}{1.1}
\caption{Ablation results on different modules.}
\vskip -2ex
\label{table3}
\begin{center}
\setlength{\tabcolsep}{2pt}
\scalebox{0.65}{
\begin{tabular}{c|ccc|c}
\hline
{Method}    & {Divider}       & {Pedestrian Crossing} & {Boundary}      & {All classes}   \\ \hline \hline
Baseline   & 40.1          & 22.2                & 39.7          & 34.0          \\
Baseline + CSL      & 43.0          & 24.7               & 42.9          & 36.9          \\
\makecell[c]{Baseline + ASL+AML }      & \textbf{43.8} & \textbf{25.7}       & \textbf{44.2} & \textbf{37.9} \\ \hline
\end{tabular}}
\end{center}
\vspace{-0.45cm} 
\end{table}
\begin{table}[tb]
\large
\renewcommand{\arraystretch}{1.1}
\caption{Ablation results on different modules.}
\vskip -2ex
\label{table4}
\begin{center}
\scalebox{0.65}{
\begin{tabular}{c|ccc|c}
\hline
Method    & Divider       &Pedestrian Crossing & Boundary      & All classes   \\ \hline
LV-teacher   & 42.8          & 24.0                & 42.6          & 36.5          \\
GV-teacher      & 42.7          & 23.6               & 43.1          & 36.5          \\
Synchronous      & 42.9          & 23.9               & 43.9          & 36.5          \\
Asynchronous      & \textbf{43.8} & \textbf{25.7}       & \textbf{44.2} & \textbf{37.9} \\ \hline
\end{tabular}}
\end{center}
\vspace{-0.85cm} 
\end{table}
\subsection{Experiment Datasets}
\textbf{nuScenes}~\cite{31} is a comprehensive and universal dataset for autonomous driving. There are six camera views in each frame which is suitable for the BEV mapping task. What is more, it has detailed intrinsic and extrinsic parameters, which are the core of the task. Based on the list of road elements in the dataset, we selected three types of road elements, namely \textit{road divider}, \textit{pedestrian crossing}, and \textit{lane divider}.

\textbf{Cam2BEV}~\cite{32} contains two synthetic datasets of semantically segmented road-scene images. There are four camera views. Furthermore, it has the ground truth in the bird's eye view. 
In this work, the road scene includes three semantic classes: \textit{road}, \textit{dynamic}, and \textit{static} objects.
\subsection{Evaluation Metrics}
Intersection over Union (IoU): The IoU between the prediction and the ground truth is given by:
\vspace{-0.15cm}
\begin{equation}
    IoU(M_{1},M_{2}) = \frac{\left|M_{1}\cap M_{2} \right|}{\left|M_{1}\cup M_{2} \right|},
    \vspace{-0.15cm}
\end{equation}
where $M_{1}$ and $M_{2}$ are the semantic prediction and the ground truth, respectively.

Chamfer Distance (CD): It is the judging criterion defined in~\cite{19}. The reference proved that it can represent a comparison of the structured outputs. It is the distance of points in each curve between the prediction and the ground truth:
\vspace{-0.15cm}
\begin{equation}
\begin{aligned}
    CD_{p}=\frac{1}{C_{1}}\sum_{x\in C_{1}} \min\limits_{y\in C_{2}}\left||x,y\right||,\\
    CD_{L}=\frac{1}{C_{2}}\sum_{x\in C_{2}} \min\limits_{y\in C_{1}}\left||x,y\right||,
\end{aligned}
\vspace{-0.15cm}
\end{equation}
where $C_{1}$ and $C_{2}$ are sets of points in the prediction and ground truth respectively. $x$ and $y$ is the pixel coordinate.
\subsection{Implementation Details}
The encoder of two streams is EfficientNet-B0~\cite{34}. The BEV decoder likes ref \cite{19}. The Adam is used to optimize the network with 30 epochs. The initial learning rate is $0.001$ and decays by a factor of $0.1$ at epochs $10$. The width and height of the BEV road map is $200{\times}400$ with respective ranges of $(-15m,15m)$ and $(-30m,30m)$, and the resolution is $0.15$.  
For the local-self view stream, the size of images in the camera coordinate system is $400{\times}800$, which is obtained by IPM. This process assumes that $y$ is $1m$ in the camera coordinate system. The $x$ and $z$ have ranges of $(-5m,5m)$ and $(3m,29m)$, respectively. That is because the learned feature maps are sized with $32 {\times} 88$.
\begin{table}[tb]
\large
\renewcommand{\arraystretch}{1.1}
\caption{Results about the start time of asynchronous mutual learning.}
\vskip -2.5ex
\label{table5}
\begin{center}
\scalebox{0.65}{
\begin{tabular}{c|ccc|c}
\hline
 Time    & Divider       & Pedestrian Crossing & Boundary      & All classes   \\ \hline
The 5th    & 43.8          & 25.7                & 44.2          & 37.9          \\
The 8th       & 42.6          & 24.4               & 42.9          & 36.6          \\
The 12th       & 40.7          & 21.9               & 40.5          & 34.4          \\ \hline
\end{tabular}}
\end{center}
\vspace{-0.35cm} 
\end{table}
\begin{table}[tb]
\large
\renewcommand{\arraystretch}{1.1}
\caption{Results about the start time of asynchronous mutual learning.
``S'' represents synchronous. ``A'' represents asynchronous.}
\vskip -2ex
\label{table6}
\begin{center}
\scalebox{0.65}{
\begin{tabular}{cc|ccc|c}
\hline
\multicolumn{2}{c|}{Variants}                       & Divider       & Pedestrian Crossing & Boundary      & All Classes   \\ \hline
L2   Loss                           & S  & 40.4          & 24.8                & 42.0          & 35.7          \\
\multirow{2}{*}{KL   Loss}          & S  & 43.2          & 24.2                & 43.7          & 37.0          \\
                                    & A & \textbf{43.6} & \textbf{24.6}       & \textbf{43.6} & \textbf{37.3} \\
\multirow{2}{*}{CE Loss} & S  & 42.9          & 23.9                & 43.9          & 36.9          \\
                                    & A & \textbf{43.8} & \textbf{25.7}       & \textbf{44.2} & \textbf{37.9} \\ \hline

\end{tabular}}
\end{center}
\vspace{-0.85cm} 
\end{table}
\subsection{Comparison against Other Methods}
We compare Bi-Mapper with four methods. The classic method is IPM~\cite{33} which is a basic component of our method. Images are learned in the pixel coordinate system. Then they are projected into the ego coordinate system by intrinsic and extrinsic parameters. Further, LSS~\cite{8} leverages depth estimation. VPN~\cite{18} and HDMapNet~\cite{19} use MLP to learn the transformation. BEVFormer~\cite{14} has a relatively good performance in the contemporary research of BEV perception. Thus, it is added to the comparison. We note that the view transformer is inconsistent in all experiments.

\textbf{Results on nuScenes:}
Table~\ref{table12} shows the comparison of results on the nuScenes dataset. Obviously, Bi-Mapper outperforms all previous methods. Besides, the running time of Bi-Mapper is comparable to that of other jobs, which is $2.4s$.
The IoU ($37.9\%$) of all classes is higher than the best performance ($35.8\%$) with $2.1\%$. 
The gain is obtained by the fusion of the proposed LV stream and GV stream, which can consider both local learning and global learning. Other approaches focus on only one of them. Then, we use the CD distance to compare the performance of different methods.
The object of CD is the semantic pixel.
$CD_{p}$ represents the precision and $CD_{L}$ denotes the recall.
It can be clearly observed that our method has outstanding performance.

\begin{table}[t]
\centering
\renewcommand{\arraystretch}{1.1}
\setlength{\tabcolsep}{4pt}
\caption{The ablation results on different weights of fusion.}
\vskip -2ex
\label{table7}
\begin{tabular}{cc|ccc|c}
\hline
\multicolumn{2}{c|}{Variants}  & \multirow{2}{*}{Divider} & \multirow{2}{*}{Pedestrian crossing} & \multirow{2}{*}{Boundary} & \multirow{2}{*}{All Classes} \\ \cline{1-2}
\multicolumn{1}{c|}{GV} & LV  &                          &                                      &                           &                              \\ \hline
\multicolumn{2}{c|}{Concat}   & 40.5                     & 25.1                                 & 41.4                      & 35.7                         \\
1.0                     & 1.0 & 37.7                     & 20.7                                 & 40.0                      & 32.8                         \\
0.1                     & 1   & 39.9                     & 21.0                                 & 41.1                      & 34.0                         \\
1                       & 0.1 & \textbf{43.8}            & \textbf{25.7}                        & \textbf{44.2}             & \textbf{37.9}                \\ \hline
\end{tabular}
\vspace{-0.3cm}
\end{table}
\begin{table}[tb]
\label{table9}
\begin{center}
\large
\renewcommand{\arraystretch}{1.1}
\setlength{\tabcolsep}{2pt}
\caption{The gain effect of LV stream including CSL and AMU.}
\vskip -0.7ex
\scalebox{0.65}{
\begin{tabular}{c|ccc|c}
\hline
Method    & Divider       &Pedestrian Crossing & Boundary      & All classes   \\ \hline
LSS    & 38.3          & 14.9                & 39.3          & 30.8          \\
\rowcolor{gray!10}LSS +    & \textbf{43.5(+5.2)}          & \textbf{24.9(+10.0)}                 & \textbf{43.7(+4.4) }         & \textbf{37.4(+6.6)}\\ \hline
HDMapNet      & 40.6          & 18.7                & 39.5          & 32.9   \\  
\rowcolor{gray!10}HDMapNet +      & \textbf{42.3(+1.7) }         & \textbf{24.4(+5.7)}               & \textbf{43.2(+3.7) }         & \textbf{36.9(+4.0) }         \\
 \hline
\end{tabular}}
\end{center}
\vspace{-0.8cm} 
\end{table}

\textbf{Results on Cam2BEV:}
Table~\ref{table8} shows the comparison on the Cam2BEV dataset.
Bi-Mapper and LSS have outstanding performances.
And Bi-Mapper is higher than LSS by $3.9\%$ in IoU.
It demonstrates that the proposed Bi-Mapper seamlessly adapts to various road scenes.

\textbf{Visualization on nuScenes:}
The visualization results are shown in Fig.~\ref{Fig.4}.
Compared with other methods, Bi-Mapper constructs highly accurate BEV road maps in various scenes. However, some faraway lines are blurry to observe.   

\textbf{Visualization on Cam2BEV:}
Fig.~\ref{Fig.6} shows the visualization of LSS and Bi-Mapper on the Cam2BEV dataset, which are comparable. Both of them have a strong ability to learn semantic information about roads and static objects. But Bi-Mapper has more advantages in identifying dynamic objects.

\subsection{Ablation Study}
\begin{figure}[t]
      \centering
      \includegraphics[scale=0.4]{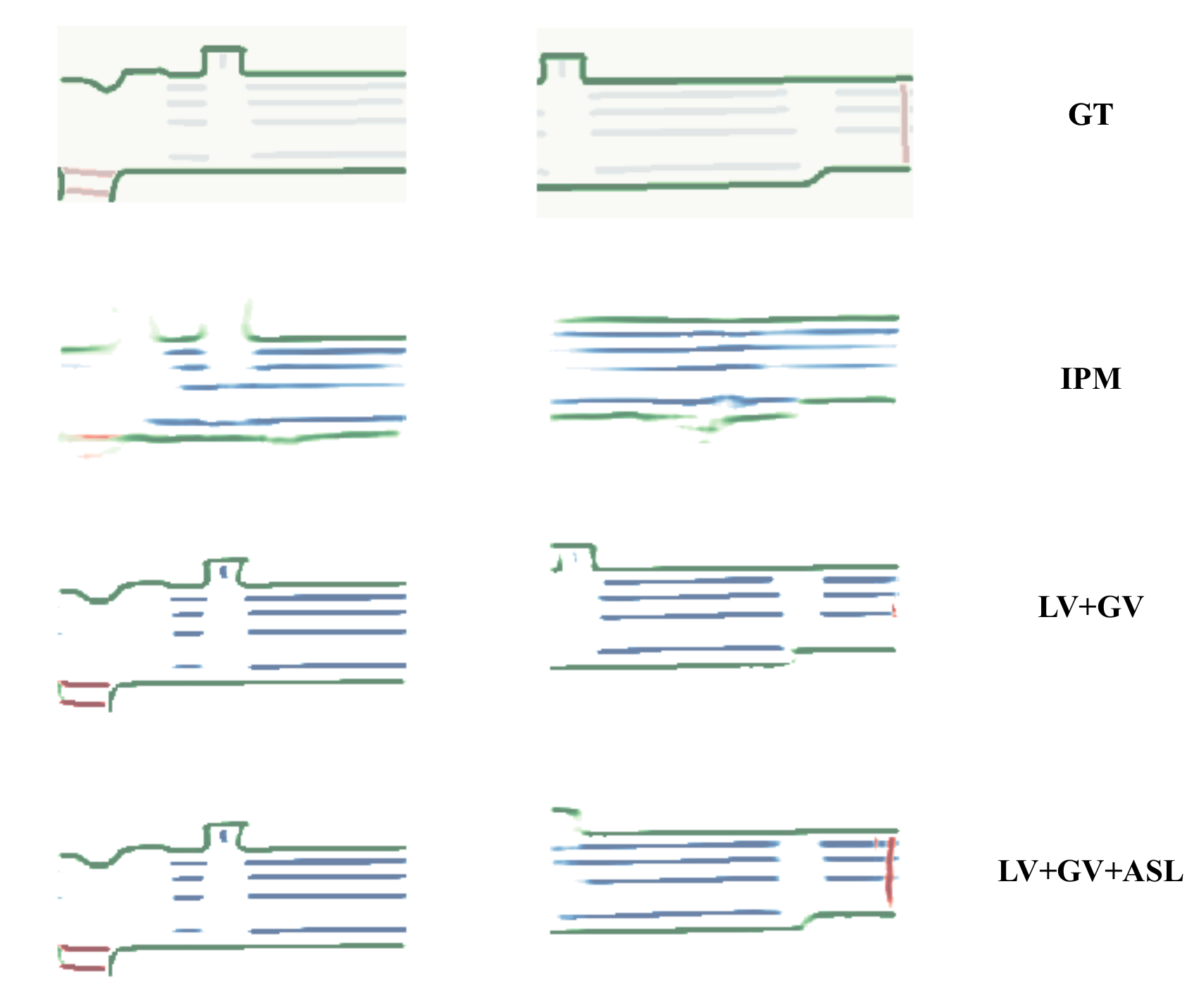}
      \vspace{-0.5cm} 
      \caption{The visual result of ablation about ASL. From the top to bottom are GT, IPM, LV+GV, and LV+GV+ASL, respectively.}
      \label{Fig.asl}
      \vspace{-0.65cm}
\end{figure}

\textbf{Core Blocks:}
The core of the BEV road map network includes two parts. One is Cross-Space (ASL), and the other is asynchronous mutual learning (AML). We first verified the effectiveness of them. The baseline includes two streams (LV and GV) which are fused in a weighted way and a segmentation loss. Then, the ASL is added. Finally, AML works in the last experiment. The results are shown in Table~\ref{table3}.
The two blocks achieve improvements with $2.9\%$ and $1.9\%$ in IoU. It demonstrates that they are beneficial for reaching robust semantic mapping. 
At the same time, we supplement visual results about the ablation experiment of ASL, as shown in Fig.~\ref{Fig.asl}. We compare three methods which are IPM, LV+GV, and LV+GV+ASL. From the results, it can be seen that IPM recognizes the distant target ambiguously. And our proposed dual-stream and ASL modules have better performance. Noticeably, these modules have a positive impact on handling the geometric distortion of IPM.

\textbf{Asynchronous Mutual Learning:} Next the effectiveness of asynchronism in mutual learning is verified, as shown in Table \ref{table4}. Four experiments are designed. First of all, it is single learning from two streams. Then synchronous and asynchronous mutual learning respectively. It proves that mutual learning is deserved. Moreover, asynchronous mutual learning outperforms synchronous mutual learning, which shows the significance of proposing asynchrony.

\textbf{Start Time:} The start time of AML is an object worth exploring, which represents the impact of mutual learning. The result is shown in Table~\ref{table5}. Note that LV-teacher starts at the beginning of training. It indicates that the start time as the fifth epoch is better. And the later asynchronous entry time will pull down the accuracy. The cause of this situation may be the long-term LV-teacher has trapped the network in a single direction of learning, with little impact from mutual learning. At the same time, we try three choices of loss, which is performed in Table~\ref{table6}.
The result of $37.9\%$ in IoU demonstrates that the proposed asynchronous mutual learning strategy is effective.

\textbf{Fusion:} Interestingly, we find that the different fusion weights will also affect the semantic mapping accuracy.
Thus, an ablation experiment about the weight difference of fusion is shown in Table~\ref{table7}.
When the weights of GV and LV are $1$ and $0.1$, it achieves the best effect.
This confirms that the LV stream still has the effects of geometric distortion and a small weight will reduce its impact.

\textbf{On the Effectiveness of Bi-Mapper:} At last, the gain effect of LV stream with ASL and AML is tested with LSS~\cite{8} and HDMapNet~\cite{19}.
Both networks will serve as alternative modules for the GV stream. 
As shown in Table~\ref{table9}, it can be found that this module can consistently bring great accuracy improvements to them. 
Therefore, it is promising that other methods with our proposed module can have a complementary performance gain.

\subsection{Application in Real Robotic Navigation Scenario}
To verify the generalization of the model, we apply our model in a real robotic navigation scenario. We collect a real-world dataset via a four-wheel transport robot in two places, including the industrial park and the campus.
It is equipped with a binocular camera and three monocular cameras.

\begin{figure}[t]
      \centering
      \includegraphics[scale=0.45]{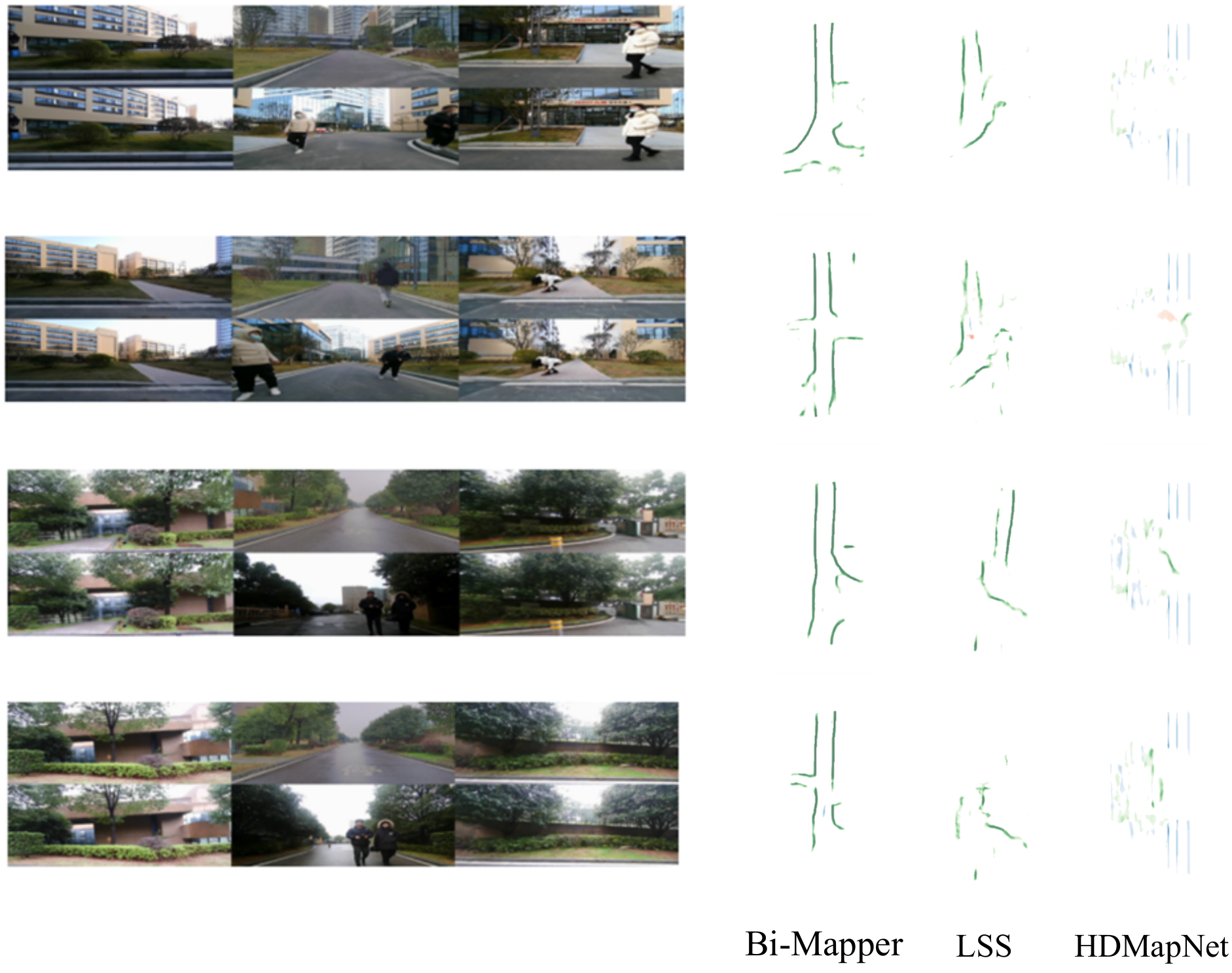}
      \vspace{-0.5cm} 
      \caption{The result in real application scenarios. The first two rows are in the industrial park and the last two rows are in the campus. Since there are only four views, the left and right images are input repeatedly.}
      \label{Fig.5}
      \vspace{-0.75cm}
\end{figure}

While there is no ground truth to measure the accuracy in these scenarios, we display a set of results visually. As shown in Fig.~\ref{Fig.5}. Although some areas are not predicted correctly, our model can output more accurate BEV road maps compared to other methods.
The reason for this case is that calibration parameters are not the same for facing different scenes. Other methods only learn the corresponding parameters from the dataset, which may not adapt to new environments.
However, our method takes calibration parameters as input attributes that are not directly learned objects.
Precisely, the IPM view after calibration parameters transformation is further learned in the LV stream. 

\section{Conclusion}
In this paper, we propose Bi-Mapper for holistic BEV semantic mapping in autonomous driving.
Bi-Mapper is a dual-stream network to construct a BEV road map from the front view and the IPM view, which adopts an across-space loss and asynchronous mutual learning to enhance top-down semantic mapping.
An extensive set of experiments on nuScenes and Cam2BEV datasets demonstrates that it has great performance and potential for consistently boosting the accuracy of various mapping methods.
Moreover, it shows robust semantic mapping results in real application scenarios.
In the future, we intend to leverage temporal information to construct more precise BEV road maps with wide ranges.

\bibliographystyle{IEEEtran}
\bibliography{bib.bib}

\end{document}